# Acquiring a Lexicon from Unsegmented Speech

Carl de Marcken
MIT Artificial Intelligence Laboratory
545 Technology Square, NE43-804
Cambridge, MA, 02139, USA
cgdemarc@ai.mit.edu

## Abstract

We present work-in-progress on the machine acquisition of a lexicon from sentences that are each an unsegmented phone sequence paired with a primitive representation of meaning. A simple exploratory algorithm is described, along with the direction of current work and a discussion of the relevance of the problem for child language acquisition and computer speech recognition.

## 1 Introduction

We are interested in how a lexicon of discrete words can be acquired from continuous speech, a problem fundamental both to child language acquisition and to the automated induction of computer speech recognition systems; see (Olivier, 1968; Wolff, 1982; Cartwright and Brent, 1994) for previous computational work in this area. For the time being, we approximate the problem as induction from phone sequences rather than acoustic pressure, and assume that learning takes place in an environment where simple semantic representations of the speech intent are available to the acquisition mechanism.

For example, we approximate the greater problem as that of learning from inputs like

Phon. Input: /ðəræbɪtsɪneʸboʷt/
Sem. Input: { BOAT A IN RABBIT THE BE }
(The rabbit's in a boat.)

where the semantic input is an unordered set of identifiers corresponding to word paradigms. Obviously the artificial pseudo-semantic representations make the problem much easier: we experiment with them as a first step, somewhere between learning language "from a radio" and providing an unambiguous textual transcription, as might be used for training a speech recognition system.

Our goal is to create a program that, after training on many such pairs, can segment a new phonetic utterance into a sequence of morpheme identifiers. Such output could be used as input to many grammar acquisition programs.

## 2 A Simple Prototype

We have implemented a simple algorithm as an exploratory effort. It maintains a single dictionary, a set of words. Each word consists of a phone sequence and a set of sememes (semantic symbols). Initially, the dictionary is empty. When presented with an utterance, the algorithm goes through the following sequence of actions:

- It attempts to cover ("parse") the utterance phones and semantic symbols with a sequence of words from the dictionary, each word offset a certain distance into the phone sequence, with words potentially overlapping.

- It then creates new words that account for uncovered portions of the utterance, and adjusts words from the parse to better fit the utterance.

- Finally, it reparses the utterance with the old dictionary and the new words, and adds the new words to the dictionary if the resulting parse covers the utterance well.

Occasionally, the program removes rarely-used words from the dictionary, and removes words which can themselves be parsed. The general operation of the program should be made clearer by the following two examples. In the first, the program starts with an empty dictionary, early in the acquisition process, and receives the simple utterance /nina/ { NINA } (a child's name). Naturally, it is unable to parse the input.

|  | Phones | Sememes |
|---|---|---|
| Utterance: | /nina/ | { NINA } |
| Words: |  |  |
| Unparsed: | /nina/ | { NINA } |
| Mismatched: |  |  |

From the unparsed portion of the sentence, the program creates a new word, /nina/ { NINA }. It then reparses

|  | Phones | Sememes |
|---|---|---|
| Utterance: | /nina/ | { NINA } |
| Words: | /nina/ | { NINA } |
| Unparsed: |  |  |
| Mismatched: |  |  |

Having successfully parsed the input, it adds the new word to the dictionary. Later in the acquisition process, it encounters the sentence *you kicked off the sock*, when the dictionary contains (among other words) /yu/ { YOU }, /ðə/ { THE }, and /rsɑk/ { SOCK }.

|  | Phones | Sememes |
|---|---|---|
| Utterance: | /yukɪktɔfðəsɑk/ | { KICK YOU OFF SOCK THE } |
| Words: | /yu/ | { YOU } |
|  | /ðə/ | { THE } |
|  | /rsɑk/ | { SOCK } |
| Unparsed: | kɪktɔf | { KICK OFF } |
| Mismatched: | r |  |

The program creates the new word /kɪktɔf/ { KICK OFF } to account for the unparsed portion of the input, and /sɑk/ { SOCK } to fix the mismatched phone. It reparses,

|  | Phones | Sememes |
|---|---|---|
| Utterance: | /yukɪktɔfðəsɑk/ | { KICK YOU OFF SOCK THE } |
| Words: | /yu/ | { YOU } |
|  | /kɪktɔf/ | { KICK OFF } |
|  | /ðə/ | { THE } |
|  | /sɑk/ | { SOCK } |
| unused | /rsɑk/ | { SOCK } |
| Unparsed: |  |  |
| Mismatched: |  |  |

On this basis, it adds /kɪktɔf/ { KICK OFF } and /sɑk/ { SOCK } to the dictionary. /rsɑk/ { SOCK }, not used in this analysis, is eventually discarded from the dictionary for lack of use. /kɪktɔf/ { KICK OFF } is later found to be parsable into two subwords, and also discarded.

One can view this procedure as a variant of the expectation-maximization (Dempster et al., 1977) procedure, with the parse of each utterance as the hidden variables. There is currently no preference for which words are used in a parse, save to minimize mismatches and unparsed portions of the input, but obviously a word grammar could be learned in conjunction with this acquisition process, and used as a disambiguation step.

## 3 Tests and Results

To test the algorithm, we used 34438 utterances from the Childes database of mothers' speech to children (MacWhinney and Snow, 1985; Suppes, 1973). These text utterances were run through a publicly available text-to-phone engine. A semantic dictionary was created by hand, in which each root word from the utterances was mapped to a corresponding sememe. Various forms of a root ("see", "saw", "seeing") all map to the same sememe, *e.g.*, SEE . Semantic representations for a given utterance are merely unordered sets of sememes generated by taking the union of the sememe for each word in the utterance. Figure 1 contains the first 6 utterances from the database.

We describe the results of a single run of the algorithm, trained on one exposure to each of the 34438 utterances, containing a total of 2158 different stems. The final dictionary contains 1182 words, where some entries are different forms of a common stem. 82 of the words in the dictionary have never been used in a good parse. We eliminate these words, leaving 1100. Figure 2 presents some entries in the final dictionary, and figure 3 presents all 21 (2%) of the dictionary entries that might be reasonably considered mistakes.

| Phones | Sememes | Phones | Sememes |
|---|---|---|---|
| /yu/ | { YOU } | /bik/ | { BEAK } |
| /ðə/ / | { THE } | /we/ | { WAY } |
| /wɑt/ | { WHAT } | /hi/ | { HEY } |
| /tu/ | { TO } | /brik/ | { BREAK } |
| /du/ | { DO } | /fɪŋgɜ/ | { FINGER } |
| /e/ | { A } | /kɪs/ | { KISS } |
| /ɪt/ | { IT } | /tɑp/ | { TOP } |
| /aɪ/ | { I } | /kɔld/ | { CALL } |
| /ɪn/ | { IN } | /ɛgz/ | { EGG } |
| /wi/ | { WE } | /θɪŋ/ | { THING } |

Figure 2: Dictionary entries. The left 10 are the 10 words used most frequently in good parses. The right 10 were selected randomly from the 1100 entries.

| | |
|---|---|
| /ɪŋ/ { BE } | /nupis/ { SNOOPY } |
| /ɪŋ/ { YOU } | /wo/ { WILL } |
| /ɪŋ/ { DO } | /zu/ { AT ZOO } |
| /ʃiz/ { SHE BE } | /don/ { DO } |
| /ʃæpɪn/ { HAPPEN } | /sɛlf/ { YOU } |
| /t/ { NOT } | /ə/ { BE } |
| /skɑtt/ { BOB SCOTT } | /smʌd/ { MUD } |
| /nidəlɪz/ { NEEDLE BE } | /ɛrɛ/ { BE } |
| /sʌmθ/ { SOMETHING } | /dont/ { DO NOT } |
| /wɑtɑrðiz/ { WHAT BE THESE } | |
| /wɑthæpɪnd/ { WHAT HAPPEN} | |
| /drɑnʌdʒwiz/ { DROWN OTHERWISE } | |

Figure 3: All of the significant dictionary errors. Some of them, like /ʃiz/ are conglomerations that should have been divided. Others, like /t/, /wo/, and /don/ demonstrate how the system compensates for the morphological irregularity of English contractions. The /ɪŋ/ problem is discussed in the text; misanalysis of the role of /ɪŋ/ also manifests itself on *something*.

The most obvious error visible in figure 3 is the suffix *-ing* (/ɪŋ/), which should be have an empty sememe set. Indeed, such a word is properly hypothesized but a special mechanism prevents semantically empty words from being added to the dictionary. Without this mechanism, the system would chance upon a new word like *ring*, /rɪŋ/, use the /ɪŋ/ {} to account for most of the sound, and build a new word /r/ { RING } to cover the rest; witness *something* in figure 3. Most other semantically-empty affixes (plural /s/ for instance) are also properly hypothesized and disallowed, but the dictionary learns multiple entries to account for them (/ɛg/ "egg" and /ɛgz/ "eggs"). The system learns synonyms ("is", "was", "am", ...) and homonyms ("read", "red"; "know", "no") without difficulty.

| Sentence | Phones | Sememes |
|---|---|---|
| this is a book. | /ðɪsɪzəbuk/ | { THIS BE A BOOK } |
| what do you see in the book? | /watduyusiɪnðəbuk/ | { WHAT DO YOU SEE IN THE BOOK } |
| how many rabbits? | /haʊmɛnirabɪts/ | { HOW MANY RABBIT } |
| how many? | /haʊmɛni/ | { HOW MANY } |
| one rabbit. | /wʌnrabɪt/ | { ONE RABBIT } |
| what is the rabbit doing? | /watɪzðərabɪtduɪŋ/ | { WHAT BE THE RABBIT DO } |

Figure 1: The first 6 utterances from the Childes database used to test the algorithm.

Removing the restriction on empty semantics, and also setting the semantics of the function words *a*, *an*, *the*, *that* and *of* to {}, the most common empty words learned are given in figure 4. The *ring* problem surfaces: among other words learned are now /k/ { CAR } and /br/ { BRING }. To fix such problems, it is obvious more constraint on morpheme order must be incorporated into the parsing process, perhaps in the form of a statistical grammar acquired simultaneously with the dictionary.

| Word | Source | Word | Source |
|---|---|---|---|
| /ɪŋ/ {} | -*ing* | /wo/ {} | ? |
| /ðə/ {} | *the* | /e/ {} | *a* |
| /o/ {} | ? | /an/ {} | *an* |
| /r/ {} | *you/your* | /əv/ {} | *of* |
| /s/ {} | plural -*s* | /z/ {} | plural -*s* |
| /i/ {} | *is*/'*s* | | |

Figure 4: The most common semantically empty words in the final dictionary.

## 4 Current Directions

The algorithm described above is extremely simple, as was the input fed to it. In particular,

- The input was phonetically oversimplified, each word pronounced the same way each time it occurred, regardless of environment. There was no phonological noise and no cross-word effects.

- The semantic representations were not only noise free and unambiguous, but corresponded directly to the words in the utterance.

To better investigate more realistic formulations of the acquisition problem, we are extending our coverage to actual phonetic transcriptions of speech, by allowing for various phonological processes and noise, and by building in probabilistic models of morphology and syntax. We are further reducing the information present in the semantic input by removing all function word symbols and merging various content symbols to encompass several word paradigms. We hope to transition to phonemic input produced by a phoneme-based speech recognizer in the near future.

Finally, we are instituting an objective test measure: rather than examining the dictionary directly, we will compare segmentation and morpheme-labeling to textual transcripts of the input speech.

## 5 Acknowledgements

This research is supported by NSF grant 9217041-ASC and ARPA under the HPCC program.